\pdfoutput=1
\documentclass[11pt]{article}
\usepackage[]{EMNLP2023}

\usepackage{times}
\usepackage{latexsym}

\usepackage[T1]{fontenc}
\usepackage[utf8]{inputenc}
\usepackage{microtype}
\usepackage{inconsolata}

\usepackage[normalem]{ulem}
\usepackage{placeins}
\usepackage{enumitem}
\usepackage{booktabs}
\usepackage{multicol}
\usepackage{multirow}
\usepackage{makecell}
\usepackage{arydshln}

\usepackage{pdfpages}

\title{Terminology-Aware Translation with Constrained Decoding \\and Large Language Model Prompting}

\author{Nikolay Bogoychev\textsuperscript{*}\qquad\qquad Pinzhen Chen\textsuperscript{*} \\
        School of Informatics, University of Edinburgh\\
        \texttt{\href{mailto:n.bogoych@ed.ac.uk}{\color{black}{n.bogoych@ed.ac.uk}}, 
        \texttt{\href{mailto:pinzhen.chen@ed.ac.uk}{\color{black}{pinzhen.chen@ed.ac.uk}}}
        }
}

\begin{document}
\maketitle
\def\thefootnote{*}\footnotetext{Equal contribution.}\def\thefootnote{\arabic{footnote}}
\begin{abstract}
Terminology correctness is important in the downstream application of machine translation, and a prevalent way to ensure this is to inject terminology constraints into a translation system. In our submission to the WMT 2023 terminology translation task, we adopt a translate-then-refine approach which can be domain-independent and requires minimal manual efforts. We annotate random source words with pseudo-terminology translations obtained from word alignment to first train a terminology-aware model. Further, we explore two post-processing methods. First, we use an alignment process to discover whether a terminology constraint has been violated, and if so, we re-decode with the violating word negatively constrained. Alternatively, we leverage a large language model to refine a hypothesis by providing it with terminology constraints. Results show that our terminology-aware model learns to incorporate terminologies effectively, and the large language model refinement process can further improve terminology recall.

\end{abstract}

\section{Introduction}

\begin{figure}[t]
    \centering
    \includegraphics[scale=0.72]{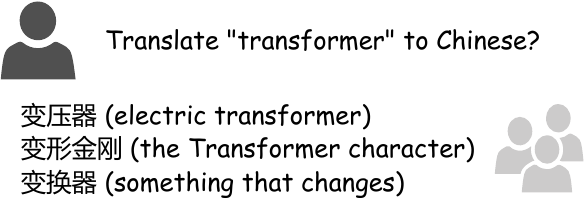}
    \caption{Terminology {hints} can help disambiguate polysemantic words when translating with limited context.}
    \label{fig:terminology_example}
\end{figure}

One of the major obstacles encountered by neural machine translation (NMT) systems pertains to the utilization of suitable domain-related words when translating specialized content not present in the training data. An illustrative instance of this challenge arises when translating ``transformer'' from English into another language, where the accurate translation depends on the context or the preference of the audience (Figure ~\ref{fig:terminology_example}). A straightforward literal translation approach often leads to suboptimal outcomes, prompting human translators unfamiliar with domain-specific knowledge to resort to reference materials for terminology precision. This issue is prevalent in the translation industry, with many commercial translation service providers offering paid solutions to address it. Furthermore, it is a popular area in machine translation research, indicated by efforts such as WMT shared tasks organization and participation focusing on terminology and domain-specific translations \citep[][inter alia]{alam-etal-2021-findings,bawden-etal-2019-findings,bawden-etal-2020-findings}.

This year's WMT terminology translation task features three language directions: German-to-English, Chinese-to-English, and English-to-Czech. In addition to reading in a source sentence, participating systems need to employ a provided dictionary, which contains source-target terminology word mappings, to incorporate into the target translation. For each source sentence in the test set, there are three modes of applying terminology constraints:
\begin{enumerate}
    \item \textit{Terminology} constraint: Dictionaries of real terminology words are provided, to be incorporated in the translations.
    \item \textit{Random} constraint: Random (but presumably correct) word mappings are obtained using a word alignment tool and provided as a pseudo-terminology dictionary.
    \item \textit{No} constraint: Source sentences can be freely translated without external information.
\end{enumerate}

We interpret that the no-constraint setting allows us to measure the competing systems' quality and understand to what degree the systems effectively utilize the provided random and terminology dictionaries. Our baseline approach is to train a terminology-aware translation (TAT) system inspired by \citet{dinu-etal-2019-training}, where, in the training data, source words are tagged with desired translations inline on the source side. Then we propose two separate refinement strategies on top of it to aggressively encourage the appearance of terminologies:
\begin{enumerate}
    \item We use a neural word aligner to identify terminology constraints missed by the baseline system, and use the same system to re-decode the source by negatively constraining (disallowing) previously incorrectly translated tokens.
    \item We also investigate the capability of a large language model to simultaneously paraphrase an existing translation to include the desired terminology constraints via curated prompts.
\end{enumerate}

Our proposed techniques can incorporate target terminology words with around 80\% recall, using automatic and soft constraints in a two-step refinement process. We observe that for German-English, our terminology-aware training and negatively constrained decoding perform better, whereas, for Chinese-English and English-Czech, LLM-based refinement achieves higher scores. In terms of overall translation accuracy, we find that negatively constrained decoding could lead to a tiny drop and LLMs are able to maintain or improve quality according to a reference-free neural metric.

\section{Related Work}
Previous research on terminology translation could be divided into two categories: \textit{soft} constraint and \textit{hard} constraint, depending on whether the resulting translation system will enforce the appearance of desired target translations. In the soft constraint setting, the convention is to train a model that is able to ingest the target terminology words inline, directly placing them after the corresponding source words in the source input \citep{dinu-etal-2019-training}. Many later implementations stem from this to include new elements such as additional lemmatization \citep{bergmanis-pinnis-2021-facilitating} or grammatical error correction \citep{pham-etal-2021-systran} as a post-processing step in order to achieve a more fluent output. Instead of placing the target constraint words inline, some other works train a system that takes the terminology constraint as either a prefix or a suffix \citep{jon-etal-2021-cuni-systems, turcan-etal-2022-constrained}.

Most hard constraint work involves post-processing a translation with desired terminologies. \citet{post-etal-2019-exploration} inserted untranslatable tokens (also known as placeholders) into the source, which will remain unchanged through the translation process. Then the placeholders are replaced with terminology words in the target language. This is entirely performed as a post-processing step. Such terminology replacement could also be done by keeping and replacing the source word at inference time, and it is also feasible to run target word replacement as post-processing \citep{molchanov-etal-2021-promt}. A hard constraint method guarantees that the chosen terminology token will appear, but often results in less fluent output, especially for morphologically rich languages because the context is not taken into consideration during replacement. It also mandates more complicated post-processing than the soft constraint approaches.

Our first post-processing proposal relies on constrained decoding, which refers to either allowing certain tokens or blocking specific tokens during inference time \citep{hokamp-liu-2017-lexically}. It has been applied to terminology injection, paraphrasing, parallel sentence mining, etc \cite{hasler-etal-2018-neural,kajiwara-2019-negative,chen-etal-2020-parallel-sentence}. We opt for negatively constraining the tokens that violated the given terminology alignments by preventing them from entering the hypothesis beam in the refinement stage. These alignments are computed using word alignment tools \citep{dyer-etal-2013-simple,dou2021word}.

Another post-processing method in our study prompts an LLM to refine a translation and incorporate terminology terms simultaneously. Whilst previous studies have explored the translation capability of LLMs \citep{vilar-etal-2023-prompting,zhang-prompting}, the works closely relevant to us are from \citet{moslem-etal-2023-adaptive} and \citet{ghazvininejad2023dictionary}. We adopt the paradigm from the latter, which re-words a constraint dictionary as a natural text and affixes it into a translation prompt. While they focused on rare words without directly benchmarking on terminology translation, our post-processing step can be seen as an extension of word-level controlled prompting to terminology translation with large language models. Both of our post-processing methods should be categorized as soft constraint approaches since there is no guarantee that negatively constrained decoding or an LLM will necessarily incorporate the constraints in a re-generation.

\section{Terminology-Aware Training}
The goal of our system implementation is to create a general-purpose terminology-aware translation system that is unsupervised and domain-agnostic, and requires the minimum effort of pre- and post-processing.

\subsection{Terminology creation}
\label{sec:terminology-aware-training}

Inspired by \citet{dinu-etal-2019-training}, we applied terminology constraints during training, but a key difference is that, unlike their approach, we assume that we have no access to downstream domain or terminology constraints during training, in order to build a general-purpose domain-agnostic system. Consequently, we have no curated terminology data to use. Therefore, we generate (pseudo-)terminology information using word alignments. Our workflow can be detailed as:
\begin{enumerate}
    \item We compute the word alignment information for the entire training set using \texttt{fast\_align} \citep{dyer-etal-2013-simple}.
    \item For each sentence, we select all bijective source-target mappings as our terminology candidates. We also filter out trivial mappings where the source and target tokens are the same (e.g. numbers, names), because those mappings are simple and hence likely to be correctly translated by a translation system even without any terminology awareness.
    \item In the training data, we replace \texttt{srcword\textsubscript{i}} in the source sentence with: \\
    \begin{tt}
        srcword\textsubscript{i} \_\_target\_\_ trgword\textsubscript{j} \_\_done\_\_
    \end{tt}
    where the $srcword\textsubscript{i}$ is the \textit{i}-th source word inside the sentence, and $trgword\textsubscript{j}$ is the word inside the target sentence, corresponding to srcword\textsubscript{i} according to word alignment information. This replacement occurs with around $10\%$ probability for each candidate source-target pair. For a sentence that does not have an associated terminology constraint, the data is the same as normal NMT.
    \item At inference time, we process the test data similarly to above, except that the source-target word mapping comes from a supplied terminology dictionary.
    
\end{enumerate}

In practice, our translation system is trained with a mix of normal translation data and terminology-injected data. The advantage of this strategy is that the trained models are general-purpose, so they can translate normal texts without terminology injection. Further, they have been exposed to a wide variety of constraints during training, making them robust to potentially unseen domain constraints.

Overall, our method is very similar to \citet{bergmanis-pinnis-2021-facilitating}'s work, except that we use whole words but not lemmas to ease pre-processing. We presume that the language model will be able to adjust the terminologies accordingly, especially for morphologically rich languages on the target side. This enables our method to be trivially transferable across languages.

Finally, our systems could easily be turned into hard-constrained by replacing the source word with the desired target terminology word. This could be feasible because our terminology-aware training installs the copying behaviour in the neural translation model, although in this mode the model would produce markedly less fluent output.

\subsection{Model architecture}
We trained Transformer-style machine translation models \citep{vaswani_trans} using the Marian NMT toolkit \citep{mariannmt}. We used the \texttt{Transformer-Big} preset which is a 6 encoder, 6 decoder architecture with 1024 hidden size, and 4096 feedforward size.\footnote{\url{https://github.com/marian-nmt/marian/blob/master/src/common/aliases.cpp\#L114}}

\subsection{Data}
The terminology task uses the same data as the constrained condition in the WMT23 general translation task. We carefully cleaned, filtered, and de-duplicated the available WMT training sets provided by the organisers, as well as the available back-translation data. After preprocessing we were left with the following:
\begin{itemize}
    \item {German-to-English} (\texttt{de-en}): 199M lines of parallel data and 29.5M lines of back-translated data.
    \item {Chinese-to-English} (\texttt{zh-en}): 21.8M lines of parallel data and 15.6M lines of back-translated data. 
    \item {Czech-to-English} (\texttt{cs-en}): 61.8M lines of parallel data and 57M lines of back-translated data.
\end{itemize}

\begin{table*}[ht]
\centering
\begin{tabular}{ll}
\toprule
\multicolumn{1}{c}{\textbf{Query}} & \multicolumn{1}{c}{\textbf{Prompt template}} \\
\midrule
\multirow{2}{*}{{Translation}} & \texttt{Source: {\$\{source\}}} \\
& \texttt{Please give me a translation in {\$\{lang\}} without any explanation.} \\
\midrule
\multirow{7}{*}{\makecell{Refinement}} & \texttt{Source: {\$\{source\}}} \\
& \texttt{Translation: {\$\{translation\}}} \\
& \texttt{Please give me a better {\$\{lang\}} translation without any explanation.} \\
& \texttt{``\$\{srcword\textsubscript{0}\}'' should be translated as ``\$\{trgword\textsubscript{0}\}'';} \\
& \texttt{``\$\{srcword\textsubscript{1}\}'' should be translated as ``\$\{trgword\textsubscript{1}\}'';} \\
& \texttt{\phantom{''}...} \\
& \texttt{``\$\{srcword\textsubscript{k}\}'' should be translated as ``\$\{trgword\textsubscript{k}\}''.} \textit{(with k >= 0)}\\
\bottomrule
\end{tabular}
\caption{Large language model prompt templates for unconstrained and constrained translation.}
\label{tab:prompts}
\end{table*}

\subsection{General quality}

The quality of our models without terminology translation is shown in Table~\ref{tab:scores}, where we report BLEU \citep{papineni-etal-2002-bleu} and COMET\textsubscript{DA}\footnote{\texttt{wmt22-comet-da}. This is a reference-based metric which requires the source input, hypothesis, and reference.} \citep{rei-etal-2020-comet} scores on test sets from the WMT22 general translation task. We note that terminology augmentation during training could result in a slight quality drop. 
\begin{table}[ht]
\centering
\begin{tabular}{ccc}
\toprule
      & BLEU           & COMET\textsubscript{DA} \\
\midrule
de-en & 31.3           & 0.8334          \\
en-cs & 39.5           & 0.8715          \\
zh-en & 20.3           & 0.7559      \\
\bottomrule
\end{tabular}
\caption{Performance of our terminology-aware translation systems in the WMT22 general translation task.}
\label{tab:scores}
\end{table}

\section{Post-Translation Terminology Injection}
Despite training our model with terminology awareness, there is no mechanism to ensure that the desired terminology constraint will appear on the target side. The neural network decoding behaviour is not entirely predictable, especially given the assumption of no additional domain adaptation. Below, we present two distinct strategies to try \textit{harder} to promote the terminology constraints, via automatic post-editing through constrained beam search and large language models.

\subsection{Negatively constrained decoding}
\label{sec:constrained-beam-search}
While it is easy enough to notice when a target terminology term is not generated as per a given constraint, it is not trivial to understand which word has been produced in place of the desired term. In order to do this, we make use of \textit{awesome-align}, a neural multilingual word aligner \citep{dou2021word}, with the following procedure:
\begin{enumerate}
    \item For each source-translation pair, we check if all required terminology terms appear on the target side. If they do, then we stop processing more rules.
    \item Then, we use \textit{awesome-align} to compute word alignments and detect the word(s) that have been generated in place of the desired terms according to the provided terminology constraints.
    \item We decode the source sentence again, penalising the words that violated the terminology constraint, by forbidding the decoder from generating them at each generation step, unless they carry more than 95\% of the probability mass at a certain step.
\end{enumerate}
In practice, this procedure can be repeated infinitely, until all terminology constraints are fulfilled, but we decided to limit it to only one iteration, to keep this a realistic production scenario in terms of computational budget.

\begin{table*}[th]
\centering\setlength{\tabcolsep}{0.8ex}
\begin{tabular}{ccccccccc}
\toprule
\multirow{2}{*}{\textbf{Mode}} & \multirow{2}{*}{\textbf{Model}} & \multirow{2}{*}{\textbf{Refine}} & \multicolumn{2}{c}{\textbf{de$\rightarrow$en}} & \multicolumn{2}{c}{\textbf{zh$\rightarrow$en}} & \multicolumn{2}{c}{\textbf{en$\rightarrow$cs}} \\
& & & \multicolumn{1}{c}{\textbf{Recall}}  & \multicolumn{1}{c}{\textbf{COMET\textsubscript{QE}}} & \multicolumn{1}{c}{\textbf{Recall}}  & \multicolumn{1}{c}{\textbf{COMET\textsubscript{QE}}} & \multicolumn{1}{c}{\textbf{Recall}}  & \multicolumn{1}{c}{\textbf{COMET\textsubscript{QE}}} \\
\midrule
\multirow{5}{*}{\makecell{\textit{terminology}\\constraints}} 
& TAT & -   & \textbf{82.30} & .0797 & 49.98 & -.0896 & 73.75 & .0601 \\
& TAT & NCD & 82.01 & .0775 & 50.42 & -.0903 & 73.26 & .0588 \\
& TAT & LLM & 64.35 & .1197 & \textbf{83.06} & \phantom{-}.0185 & 76.00 & .0866 \\
& LLM & -   & 41.86 & .1244 & 46.63 & \phantom{-}.0191 & 48.14 & .0913 \\
& LLM & LLM & 70.48 & .1180 & 81.01 & \phantom{-}.0201 & \textbf{78.94} & .0882 \\
\hdashline
\multirow{4}{*}{\makecell{\textit{no}\\constraint\textsuperscript{$\dagger$}}} 
& TAT & -   & 39.82 & .1085 & 13.64 & -.1163 & 48.11 & .0712 \\
& TAT & LLM & 39.59 & .1251 & 42.76 & \phantom{-}.0203 & 47.31 & \textbf{.0955} \\
& LLM & -   & 41.86 & .1244 & 46.63 & \phantom{-}.0191 & 48.14 & .0913 \\
& LLM & LLM & 39.65 & \textbf{.1258} & 46.72 & \textbf{\phantom{-}.0228} & 46.22 & .0943 \\
\midrule
\multirow{5}{*}{\makecell{\textit{random}\\constraints}} 
& TAT & -   & \textbf{76.17} & .0716  & 81.55 & -.1105 & 57.10 & .0502 \\
& TAT & NCD & 75.79 & .0698  & \textbf{82.03} & -.1123 & 56.42 & .0465 \\
& TAT & LLM & 61.46 & .1206  & 63.17 & \phantom{-}.0175 & 70.97 & .0875 \\
& LLM & -   & 38.70 & .1244  & 52.49 & \phantom{-}.0191 & 39.34 & .0913 \\
& LLM & LLM & 66.74 & .1188  & 67.10 & \phantom{-}.0196 & \textbf{73.37} & .0867 \\
\hdashline
\multirow{4}{*}{\makecell{\textit{no}\\constraint\textsuperscript{$\ddagger$}}} 
& TAT & -   & 35.60 & .1085 & 36.18 & -.1163           & 37.35 & .0712 \\
& TAT & LLM & 37.58 & .1251 & 49.48 & \phantom{-}.0203 & 39.03 & \textbf{.0955} \\
& LLM & -   & 38.70 & .1244  & 52.49 & \phantom{-}.0191 & 39.34 & .0913 \\
& LLM & LLM & 37.62 & \textbf{.1258} & 49.00 & \textbf{\phantom{-}.0228} & 38.42 & .0943 \\
\bottomrule
\multicolumn{9}{l}{\textsuperscript{$\dagger$}\small{Recall computed against terminology constraints.}} \\
\multicolumn{9}{l}{\textsuperscript{$\ddagger$}\small{Recall computed against random constraints.}} \\
\end{tabular}
\caption{Terminology recall and translation quality measured by COMET\textsubscript{QE} of our systems on the \textit{blind test} set. \\
TAT: terminology-aware translation; NCD: negatively constrained decoding; LLM: large language model.}
\label{tab:results-test}
\end{table*}

\subsection{Large language models}
Recent years saw the rise of large language models (LLMs), which have a strong capability in various NLP tasks. In this paper, we investigate the effectiveness of using a large language model to generate terminology terms during translation by adding constraints to \citet{Chen2023IterativeTR}'s translation refinement prompts. We use two distinct prompts: free translation and translation refinement queries. The translation query sends a source sentence and requests a translation in the target language without any other information. On the other hand, the refinement query feeds back an unconstrained translation together with terminology constraints to request a new translation. This essentially forms an LLM version of the constrained beam search discussed in Section~\ref{sec:constrained-beam-search}. The constraints are enforced through natural language instructions in the prompts, under the situation where the softmax distribution from an LLM is not accessible by users.

The LLM we use is OpenAI's GPT-3.5.\footnote{\texttt{gpt-3.5-turbo-0613}, a snapshot of the GPT-3.5 model on 13 June 2023} It is a closed-source commercial system, where the model weights and the inference states are not available to users. The model has a context window of 4096 which is sufficient to cover an instruction, a source sentence, several terminology constraints, as well as the target translation. It is public to all users at a relatively cheap cost. In our settings, each translation is carried out in a new query session.

In Table~\ref{tab:prompts} we outline the two prompt templates we used. During querying, the placeholder variables are substituted with corresponding string values. For the refinement query, when a terminology dictionary is supplied, the source and target words are fed to the LLM via the prompt \citep{ghazvininejad2023dictionary}; if there is no terminology dictionary, the query simply asks for a refined translation. The two-step experiment with LLMs can be summarized as follows:
\begin{enumerate}
    \item We obtain an initial unconstrained translation, which may or may not fulfil all the terminology constraints. It can come from either the LLM itself or the terminology-aware translation model built in Section~\ref{sec:terminology-aware-training}.
    \item We query the LLM with the constrained translation prompt to obtain a refined translation with terminology incorporated in the prompt.
\end{enumerate}

\section{Results and Discussions}
We present our \textit{blind test} results in Table~\ref{tab:results-test}, which include both terminology recall and COMET\textsubscript{QE} scores computed by us.\footnote{\texttt{wmt21-comet-da-qe}} We used COMET\textsubscript{QE} in particular because it does not require references which are not accessible to us. We assess the effectiveness of our methods by comparing the terminology recall of our systems with and without applying terminology constraints, in both \textit{random} and \textit{real terminology} scenarios.

\subsection{Translation quality}
In terms of translation quality reflected in COMET\textsubscript{QE}, we observe that the LLM rows attain superior results, which is not surprising considering that we use an unconstrained commercial model GPT-3.5. By comparing TAT with TAT+NCD, or comparing LLM with LLM+LLM under a constrained scenario, we conclude that applying terminology constraints usually lead to a sacrifice in translation quality regardless of the language direction or the systems involved. Nonetheless, as a contrasting experiment with no constraint, LLM+LLM achieves a slightly better COMET\textsubscript{QE} score than using an LLM to translate without refinement.

Our model performed poorly on the \texttt{zh-en} task in terms of COMET\textsubscript{QE} scores. We suspect that this is because of the domain mismatch between the translation data from the general domain and the Chinese terminology test set. Upon manual inspection, we found that the latter includes web novels and literal writing which are likely to be under-represented in the generic training data.

\subsection{Terminology recall}
Focusing on terminology generation, compared with TAT or LLM in unconstrained settings, TAT marks 30-40 higher recall of terminology terms in the constrained \textit{terminology} and \textit{random} settings. This indicates that our terminology-aware training is effective in teaching translation models to follow customized source-target word alignments.

Next, as a post-processing step, negatively constrained decoding seems to be disappointing in practice. TAT+NCD often produces worse results than TAT alone in terms of both quality and terminology recall, except for \texttt{zh-en} with \textit{random} constraints. We hypothesize that this could be due to two problems: (1) word alignment errors could propagate into this process, and (2) by applying NCD, we might capture a missed terminology term but at the cost of mis-translating other words. Our constraining procedure might be improved by performing shortlisting, namely positively constrained decoding, as opposed to negatively limiting the beam search in an iterative approach.

We find the results promising when using LLMs for terminology injection. Looking at LLM+LLM versus LLM alone in various constrained conditions, terminology recall improves significantly with very little drop in overall quality. Also by comparing TAT+LLM with TAT alone, we observe that TAT and LLMs each have their own merits depending on the language direction. In terms of recall, TAT wins in \texttt{de-en}, TAT+LLM wins in \texttt{zh-en}, and they are close in \texttt{en-cs}. However, TAT+LLM is way ahead if measured by COMET\textsubscript{QE}. However, we must note that an LLM costs significantly more resources than a dedicated translation model at both training and inference time.

\FloatBarrier
\section{Conclusion and Future Work}
We participated in all tracks of the WMT 2023 terminology shared task with a terminology-aware translation baseline, and two distinct refinement procedures using negatively constrained beam search and large language models separately. The results we produced gave us insights into the pros and cons of our systems. In future work, we could explicitly enforce the generation of the terminology token by identifying the appropriate time step and manipulating the probability distribution after softmax computation, even in an open-source large language model. This is not entirely trivial due to the presence of subwords but could be achievable.

\section*{Acknowledgement}
This project has received funding from UK Research and Innovation (UKRI) under the UK government’s Horizon Europe funding guarantee [grant numbers 10052546 and 10039436].

\bibliography{custom}
\bibliographystyle{acl_natbib}

\end{document}